%% file: main.tex
\pdfoutput=1
\documentclass[10pt,twocolumn,letterpaper]{article}

\usepackage[T1]{fontenc}

\usepackage[pagenumbers]{paperstyle} %

\input{preamble}
\usepackage{booktabs}
\usepackage[dvipsnames]{xcolor}  %
\usepackage{graphicx}
\graphicspath{{figures/}}

\definecolor{linkblue}{rgb}{0.21,0.49,0.74}
\usepackage[pagebackref,breaklinks,colorlinks,allcolors=linkblue]{hyperref}

\title{SetCon: Setup-Conditioned IMU-Language Learning from 3D Human Motion}  %

\author{Lennart Schiweck$^{1}$ \qquad Alaa Saleh$^{2}$ \qquad Crist\'obal Curio$^{1}$\\
$^{1}$Cognitive Systems Group, Reutlingen University, Germany \qquad $^{2}$Bosch Sensortec GmbH, Germany}

\begin{document}
\twocolumn[{%
\renewcommand\twocolumn[1][]{#1}%
\maketitle
\input{floats/fig1}
}]

\input{sec/0_abstract}
\input{sec/1_intro}
\input{sec/2_related}

\input{sec/3_method}

\input{sec/4_exp_a}
\input{sec/4_lang}   %
\input{sec/4_exp_b}
\input{sec/5_conclusion}

\section*{Acknowledgments}
This research was funded by the European Regional Development Fund (ERDF) within the project `AIDA' (grant FEIH 2476184). L.~Schiweck was supported by a scholarship from Bosch Sensortec GmbH. Views and opinions expressed are those of the authors only and do not necessarily reflect those of the European Union. Neither the European Union nor the granting authority can be held responsible for them.

The authors used writing and coding assistants (Claude by Anthropic, ChatGPT by OpenAI) to draft and edit text, captions, tables and figure scripts, to write and run experiment code and to search related work. The authors verified every number, claim and reference and take full responsibility for the content.

{
    \small
    \bibliographystyle{ieeenat_fullname}
    \bibliography{references}
}

\end{document}

%% file: preamble.tex
\input{macros}

%% file: floats/fig1.tex
\begin{center}
\centering
\vspace{-1.2em}%
\includegraphics[width=\textwidth]{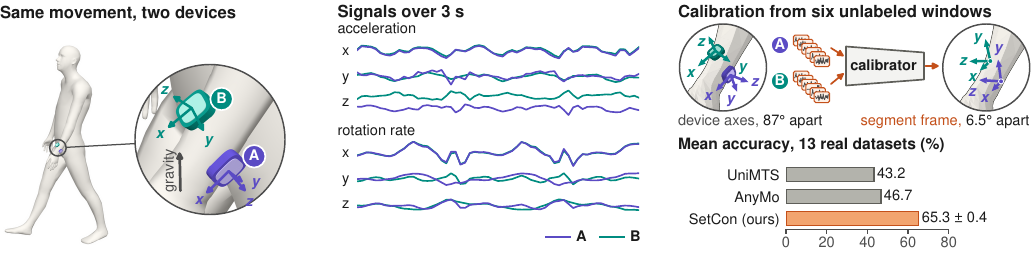}%
\captionof{figure}{\textbf{Same movement, two devices.} Two virtual devices on one forearm (left), \pv[1]{fig1.dist}\,cm apart and rotated by \pv[0]{fig1.mount}$^\circ$ against each other, record one Nymeria~\cite{ma2024nymeria} walk differently. For this pair, chosen for illustration, the axes of the two devices are \pv[1]{fig1.axes.after}$^\circ$ apart in the forearm's segment frame (right) once SetCon's calibrator has estimated each device's setup reference from six unlabeled windows. The bars give the mean accuracy of UniMTS~\cite{zhang2024unimts}, AnyMo~\cite{chen2026anymo} and SetCon with anchors on \pv[0]{set.m13.n} real datasets (\cref{tab:main}, last row, $\pm$ sd over three seeds).}%
\label{fig:teaser}%
\end{center}

%% file: sec/0_abstract.tex
\begin{abstract}
Aligning inertial measurement unit (IMU) signals with text enables zero-shot activity recognition across wearable devices.
A worn device's signal depends on its setup, the combination of body site and mounting.
Leading IMU-language models make their encoders invariant to the mounting, which also discards the device's orientation relative to the body, such as the direction of gravity that separates sitting, standing and lying.
We introduce SetCon, which replaces full invariance with selective invariance.
It estimates each device's mounting on its known body site from a few unlabeled windows and conditions its encoder on this estimate.
Because the mounting stays nearly fixed while the device is worn, the moving body serves as a calibration target.
Both calibration and representation learning are supervised by 3D human motion, since one movement can be simulated under many known sensor setups.
At test time, a class is represented by its name or by an anchor, which averages the text embeddings of the name and of activity descriptions and needs no labeled window.
A language model built on the frozen encoder also answers how and in what order a movement is performed.
Trained only on IMU simulated from the same Nymeria motion capture as AnyMo, SetCon reaches \pv[1]{mix.psel6.w12.nm.acc.m13}\,\% mean zero-shot accuracy on \pv[0]{set.m13.n} real datasets with class names and \pv[1]{mix.psel6.w12.ac.acc.m13}\,\% with anchors, against \pv[1]{anymo.window.w12.nm.acc.m13}\,\% and \pv[1]{anymo.window.w12.ac.acc.m13}\,\% for AnyMo.
\end{abstract}

%% file: sec/1_intro.tex
\section{Introduction}
\label{sec:intro}

An inertial measurement unit (IMU) in a smartwatch, a phone or any other wearable records how the body segment it sits on accelerates and rotates. %
If an encoder mapped \pvc{10}-s windows of this signal into the embedding space of a text encoder~\cite{moon2023imu2clip,girdhar2023imagebind}, as CLIP~\cite{radford2021clip} does for images, each window could be classified zero-shot as the activity whose text embedding lies nearest, on any wearable and without labeled data.
Because real IMU with activity labels is scarce, recent encoders learn from virtual IMU~\cite{kwon2020imutube,leng2023imugpt,leng2024imugpt2}, simulated on 3D human body models driven by recorded motion~\cite{zhang2024unimts,chen2026anymo}.
UniMTS~\cite{zhang2024unimts} and AnyMo~\cite{chen2026anymo} align such virtual IMU with text and classify activities on real datasets from class-name prompts.
We evaluate on the \pv[0]{set.m13.n} real datasets of the UniMTS suite that record a gyroscope on every device, so that UniMTS, AnyMo and our model each receive their full input. On Mean-13, their mean accuracy, UniMTS and AnyMo reach \pv[1]{unimts.window.w12.nm.acc.m13} and \pv[1]{anymo.window.w12.nm.acc.m13}\,\% with class names. %

The obstacle is the sensor setup~\cite{kunze2008displacement,kunze2014placement,yurtman2017orientation}, which consists of the body site where a device is worn, such as a wrist or the hip, and its mounting, how it is attached and rotated there. %
Because the same movement produces different signals under different mountings (\cref{fig:teaser}), UniMTS and AnyMo make their encoders invariant to the mounting. UniMTS uses random rotation augmentation and AnyMo learns motion tokens that do not depend on how a device is placed within its body segment.
Invariance removes the mounting but also the direction of gravity relative to the body.
For a motionless wearer, this direction indicates whether trunk and thighs are upright or horizontal and thus separates sitting, standing and lying.
A hip device worn upright on a standing person and the same device rotated by \pvc{90}$^\circ$ on a person lying on the back record the same window, with gravity along the same device axis and zero angular velocity. %
Accordingly, on the \pv[0]{post.ndatasets} test datasets with at least 20 windows of each posture, UniMTS queried with anchors classifies on average \pv[0]{post.unimts.ac.stand}\,\% of the sitting, standing and lying windows as standing, while AnyMo classifies \pv[0]{post.anymo.ac.sit}\,\% of them as sitting. %

\input{floats/fig2}Most deployments, however, provide the information that these models leave unused. %
The body site of a device is known and its mounting stays nearly fixed while the device is worn, so the windows it records during that time share one mounting. %
A few of these windows, unlabeled and drawn at random, suffice to estimate how the device is oriented on its segment, because the moving body serves as a calibration target~\cite{kunze2009facing,zuo2025tic}. %
We therefore introduce SetCon (\cref{fig:method}, top), which estimates the setup and provides it to its encoder instead of removing it. %
A calibrator estimates this orientation and its uncertainty from the body site and unlabeled windows of the device as the \emph{setup reference} (\cref{sec:reference}), the rotation from device axes into the \emph{segment frame}, axes fixed to the body segment (\cref{fig:teaser}). %
The encoder receives every window rotated so that gravity points up~\cite{mizell2003gravity,mahony2008complementary}, which keeps the vertical axis explicit and with it the direction in which the wearer turns.
It also receives each window's gravity in the segment frame, which distinguishes the two hip devices. %

3D human motion makes this conditioning learnable.
A device simulated on the 3D body model that defines the segment frame has a known mounting, which gives the calibrator its target and a measure of its own errors.
One movement can be simulated under many setups, so that the encoder learns how it appears under each, whereas a real recording carries one mounting per device that datasets rarely record. %
Like prior work~\cite{zhang2024unimts,chen2026anymo}, SetCon generates its training data from 3D motion, but also uses it to supervise the setup estimate. %

The setup reference conditions the IMU side of the alignment.
On the text side, the training captions and the class prompts determine how much of the separation in the embeddings carries over to zero-shot classification. %
SetCon trains on both the human narrations of the motion capture~\cite{ma2024nymeria} and captions that describe kinematics, generated by a large language model (LLM) from the measured 3D motion of the same windows. %
At test, a class is represented by the text embedding of its name or by an anchor, a CLIP-style prompt ensemble that averages the text embeddings of the name, its nearest training narrations and LLM-written descriptions of the activity~\cite{pratt2023cupl,menon2023description}, so no labeled window is needed. %
Because a class name does not say how or in what order a movement is performed, a small language model built on the frozen encoder~\cite{liu2023llava} is trained to answer such questions.

Our contributions are as follows.
{%
\begin{itemize}%
\item We propose setup conditioning for IMU-language learning, which replaces full invariance with \emph{selective invariance}. The sensor setup becomes an input, so that the encoder can discount how a device is mounted while it keeps the direction of gravity relative to the body and the direction in which the wearer turns. A calibrator estimates the setup from a few unlabeled windows of the device, which it selects by its tilt uncertainty (\cref{sec:reference,sec:test}). %
\item We use existing 3D human motion to supervise the setup estimate of an IMU-language encoder, not only to generate training data. Real IMU datasets rarely record the mounting, whereas a simulated device has a known one. Simulation therefore provides the calibrator's target and lets the encoder learn how one movement appears under many setups (\cref{sec:virtual,sec:reference,sec:stage1}).
\item We extend language access to the encoder beyond class names. For training, an LLM generates captions that describe kinematics from the measured 3D motion. At test, anchors need no labeled window and improve every model, most of all when the encoder is trained with these captions. A language model on the frozen encoder is more accurate than class-name prompts and answers how and in what order a movement is performed (\cref{sec:annotation,sec:test,sec:lmbridge,sec:language}). %
\item Trained only on simulated IMU, SetCon achieves state-of-the-art zero-shot accuracy on average over \pv[0]{set.m13.n} real datasets, with class names and with anchors. Its embeddings also separate the activities best, measured without any text prompt (\cref{sec:main}). %
\end{itemize}}

%% file: floats/fig2.tex
\begin{figure*}[t]
\centering
\includegraphics[width=\textwidth]{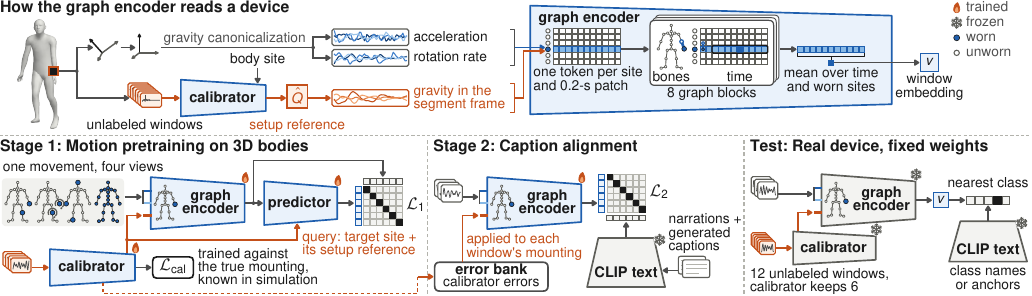}%
\caption{\textbf{Overview of SetCon.} The top row shows how the graph encoder, a spatio-temporal graph network over the body sites, reads a device. Every sample is rotated so that gravity points up and a calibrator estimates the setup reference $\hat Q$ (orange) from unlabeled windows and the body site, so that the encoder also receives gravity in the segment frame. In the bottom row, stage~1 learns how a movement appears under a target setup and stage~2 aligns the encoder with captions under calibrator errors from an error bank. The weights stay fixed at test.}
\label{fig:method}
\end{figure*}

%% file: sec/2_related.tex
\section{Related work}
\label{sec:related}

\paragraph{IMU-language pretraining.}
IMU2CLIP~\cite{moon2023imu2clip} and ImageBind~\cite{girdhar2023imagebind} bind head-worn IMU to egocentric video and text. UniMTS~\cite{zhang2024unimts} and AnyMo~\cite{chen2026anymo} align IMU simulated on body models with text from an LLM, while ZeroHAR~\cite{roychowdhury2025zerohar}, LanHAR~\cite{yan2025lanhar}, IMUZero~\cite{su2025imuzero} and HALO~\cite{ding2026halo} add sensor or activity descriptions to training on labeled real IMU.
ZeroHAR, HALO and AnyMo describe each sensor in text, whereas SetCon estimates from unlabeled windows how a test device is mounted on its known site. %

\paragraph{Virtual IMU from body models.}
Inertial pose estimation, from DIP to MobilePoser~\cite{huang2018dip,yi2021transpose,yi2022pip,mollyn2023imuposer,xu2024mobileposer}, trains on IMU synthesized from AMASS~\cite{mahmood2019amass} on SMPL bodies~\cite{loper2015smpl}. IMUTube~\cite{kwon2020imutube} and IMUGPT~\cite{leng2023imugpt,leng2024imugpt2} derive virtual IMU from video and text-to-motion. WIMUSim~\cite{oishi2025wimusim}, W2W~\cite{ray2025w2w} and IMUCoCo~\cite{zhou2025imucoco} model placement and device variation.
SetCon simulates its devices on Momentum Human Rig (MHR)~\cite{ferguson2025mhr} fits of Nymeria~\cite{ma2024nymeria,detone2026nymeriaplus} and uses the body model not only to generate data and to vary the placement, as these works do, but also to supervise mounting estimation. %

\paragraph{Placement, orientation and test-time adaptation.}
Displacement, placement and facing are long-standing problems in on-body recognition~\cite{kunze2008displacement,kunze2014placement}, answered by invariance through orientation-removing transforms~\cite{yurtman2017orientation}, rotation-invariant backbones~\cite{baek2026trihar} or augmentation~\cite{zhang2024unimts}, or by alignment with gravity through averaging~\cite{mizell2003gravity} or attitude filters~\cite{mahony2008complementary} and with the gait direction~\cite{kunze2009facing}.
The Transformer IMU Calibrator~\cite{zuo2025tic}, also trained with simulated offsets, estimates sensor-to-bone offsets from short windows for inertial motion capture and is closest to our calibrator.
Test-time adaptation updates weights or statistics on unlabeled test data~\cite{sun2020ttt,wang2021tent,zhao2025tartanimu}.
SetCon changes no weights at test and does not rotate the signal into the segment frame, whose forward axis is ambiguous under gait. An encoder pretrained to use the estimate receives it with the signal. %

\paragraph{Descriptions and language models for IMU.}
Pratt~\etal~\cite{pratt2023cupl} and Menon and Vondrick~\cite{menon2023description} replace class-name prompts with LLM-written descriptions, while Ghosh~\cite{ghosh2026modalitygap} writes them by hand for sensor prototypes. HARGPT~\cite{ji2024hargpt}, LLaSA~\cite{shouborno2025llasa}, SensorLLM~\cite{li2025sensorllm}, EgoLM~\cite{hong2025egolm}, Ego4o~\cite{wang2025ego4o} and Mojito~\cite{shan2025mojito} feed IMU or egocentric motion signals into LLMs. Zhang~\etal~\cite{zhang2024vlmclassify} find that vision-language models classify images worse than the CLIP image encoder they are built on~\cite{radford2021clip}.
SetCon builds anchors from such descriptions for its frozen IMU embeddings and trains a language model on top of its encoder as in LLaVA~\cite{liu2023llava}.

%% file: sec/3_method.tex
\section{Method}
\label{sec:method}

SetCon is trained on virtual IMU from 3D bodies and applied to real IMU with fixed weights (\cref{fig:method}).
For every device, virtual or real, a calibrator estimates the setup reference from its unlabeled windows and the encoder receives it with each window rotated so that gravity points up. %
Stage~1 pretrains the encoder on movements simulated under several views (sets of worn devices with their setups) and stage~2 aligns it with captions under calibrator errors. %

\subsection{Virtual devices on 3D bodies}
\label{sec:virtual}

A virtual device sits on a body site with a mounting, given by a skin point and a rotation $M_d$ from joint to device axes. This mounting is known for every simulated device but rarely recorded for a real one. %
We train on Nymeria's motion capture~\cite{ma2024nymeria}, represented as MHR bodies~\cite{ferguson2025mhr} and cut into consecutive, non-overlapping windows of up to \pvc{10}\,s. Nymeria also provides wristband IMU and narrations. %
Stage~1 uses all \pv[0]{data.s1.train.windows} training windows and stage~2 the \pv[0]{data.d3.train.windows} of them that overlap a narration. %
As in UniMTS~\cite{zhang2024unimts}, signals are sampled at \pvc{20}\,Hz on a graph of \pvc{22} sites that take position and axes from MHR joints. Every site is a node of the encoder, worn or not. %
Each site has eight skin points on the MHR surface and each wrist has thirty on a ring around the distal forearm. In half of the windows the wrist device follows the hand and in the other half only the pronation of the forearm. %
In stage~1, $M_d$ follows the skin surface as in AnyMo~\cite{chen2026anymo} and is drawn as $M_d=R_y(\epsilon_2)\,R_x(\epsilon_1)\,R_z(\psi)\,T_d^{\top}$, where the columns of $T_d$ are two tangents and the outward normal of the skin at the device, $R_x$, $R_y$ and $R_z$ rotate about the coordinate axes, $\psi$ is a uniform rotation about the normal and the tilts $\epsilon_1$ and $\epsilon_2$ are uniform within $\pm\pvc{10}^\circ$. %
Re-wearing a device draws a new skin point and a new $M_d$.
The accelerometer reads the specific force $f_d$ (the acceleration minus gravity in device axes) and the gyroscope reads the angular velocity $\omega_d$, both computed by central differences over the sample interval $\Delta t$ as %
\begin{equation}
\begin{aligned}
f_d(t)&=M_dR_d(t)^{\top}\big(\ddot p_d(t)-g\big),\\
\omega_d(t)&=\frac{M_dR_d(t)^{\top}}{2\Delta t}\,\log\!\big(R_d(t{+}\Delta t)\,R_d(t{-}\Delta t)^{\top}\big)^{\vee},
\end{aligned}
\label{eq:imu}
\end{equation}
where $R_d$ is the world orientation of the joint, $p_d=J_d+R_d\,r_d$ the device position given by the joint position $J_d$ and the skin point $r_d$ in joint axes, $g$ gravity, $\log$ the matrix logarithm and $(\cdot)^{\vee}$ the vector of a skew-symmetric matrix.
We then apply an \pvc{8}-Hz low-pass filter and add white noise and a per-window constant bias. %
On \pv[0]{val.nym.devices} wrists of \pv[0]{val.nym.sessions} held-out recordings, virtual magnitudes and those of Nymeria's real wristband correlate per window with a median of \pv[2]{val.nym.corr.acc} for the specific force and \pv[2]{val.nym.corr.gyro} for the angular velocity, compared with at most \pv[2]{val.nym.corr.acc.wrong} and \pv[2]{val.nym.corr.gyro.wrong} when a window is paired with another window, its time reverse or the other wrist. %

\subsection{Setup reference and gravity canonicalization}
\label{sec:reference}

An accelerometer senses gravity, so the signal of every device contains its tilt. This tilt has two sources. The tilt of the body segment separates postures, whereas the way the device is mounted on it is arbitrary. Gravity therefore reaches the encoder by two routes (\cref{fig:method}, top row). Gravity canonicalization removes the tilt from the motion signals. The setup reference returns the segment's part. %

\paragraph{Gravity canonicalization.}
Every sample is rotated so that gravity points up, by the shortest rotation $R_\uparrow(t)$ that turns the estimated gravity direction $\hat g_d(t)$ in device axes onto the vertical axis.
A complementary filter~\cite{mahony2008complementary} with gain \pvc{1}\,s$^{-1}$ estimates $\hat g_d$ from accelerometer and gyroscope. On accelerometer-only devices a zero-phase \pvc{0.3}-Hz low-pass of the accelerometer replaces it. %
Because one rotation acts on both signals, the signed rotation rate about the vertical is preserved and distinguishes left from right turns.
Canonicalization cannot, however, separate device tilt from segment tilt, so it removes both.

\paragraph{Setup reference.}
The reference describes how a device sits on its body segment (\cref{fig:teaser}). On the 3D body model, each site has a segment frame $W_d$, measured once in joint axes on \pvc{60} simulated Nymeria recordings. We use the quarter of their \mbox{10-s} windows with the most rotation, where the body is in motion. The up axis of $W_d$ is the mean vertical direction in these windows and its forward axis the principal horizontal axis of their dynamic acceleration. %
The reference is the fixed rotation from device axes into $W_d$, which covers the mounting's rotation but not its skin point. For a simulated device it is known, $Q^*_d=W_dM_d^{\top}$.
Applied to the gravity direction, it yields gravity in the segment frame, the tilt of the segment whatever the mounting.

\paragraph{Calibrator.}
A real device does not report its mounting. The calibrator $C_\phi$ therefore estimates the reference from $k$ unlabeled windows $X_{d,1},\dots,X_{d,k}$ of one worn device, its \emph{context}, raw in device axes with gravity included, together with a missing-gyroscope flag and the region and side $\ell_d$ of the site.
Two cues make this possible. Because the up axis of $W_d$ is the mean vertical in motion, gravity averaged over a few windows determines it in device axes. Gait fixes the forward axis, but only up to its sign, so $Q^*_d$ and $R_y(\pi)Q^*_d$ are equally valid targets, where $R_y(\pi)$ is the rotation by $\pi$ about the up axis of $W_d$, its $y$ axis.

A 1-D convolutional network encodes each window, with embeddings of $\ell_d$ and the flag added, into a feature $w_n$. A multilayer perceptron (MLP) head on the order-invariant pooling $[\operatorname{mean}_nw_n;\max_nw_n]$ outputs a density $p_\phi(Q)$ over rotations, a mixture of two Matrix-Fisher components on $SO(\pvc{3})$, so that the estimate carries its own uncertainty. %
The estimate $\hat Q_d=C_\phi\big(\{X_{d,n}\}_{n=1}^{k},\ell_d\big)$ is the mode of the heavier component. Its tilt uncertainty $\sigma_\uparrow$ is the spread of that component's up axis. %
The calibration loss is the negative log-likelihood of the target, symmetrized over $R_y(\pi)$, plus an equivariance term $\mathcal L_{\mathrm{eq}}$,
\begin{equation}
\mathcal L_{\mathrm{cal}}=-\log\tfrac{1}{2}\big[p_\phi(Q^*_d)+p_\phi(R_y(\pi)\,Q^*_d)\big]+\lambda_{\mathrm{eq}}\,\mathcal L_{\mathrm{eq}},
\label{eq:cal}
\end{equation}
where $\lambda_{\mathrm{eq}}=\pv{impl.s1.lameq}$. The term $\mathcal L_{\mathrm{eq}}$ requires that rotating all raw vectors of the context by a uniformly random $S\in SO(\pvc{3})$ turns the predicted density with them, from $Q$ to $QS^{\top}$. %
The target $Q^*_d$ requires the true mounting, which only a simulated device provides. The calibrator is therefore pretrained with this loss on simulated devices, refined in stage~1, then frozen.

\paragraph{Encoder input.}
The nine channels of the encoder input hold the canonicalized specific force and angular velocity (acceleration and rotation rate in \cref{fig:method}) divided by fixed scales $\kappa_f$ and $\kappa_\omega$, followed by the reference channels $b_d=\hat Q_d\hat g_d$, gravity in the segment frame,
\begin{equation}
x_d(t)=\big[R_{\uparrow}f_d/\kappa_f;\ R_{\uparrow}\,\omega_d/\kappa_\omega;\ \hat Q_d\,\hat g_d\big](t),
\label{eq:input}
\end{equation}
where $R_\uparrow=R_\uparrow(t)$. The scales $\kappa_f=\pv[2]{stats.mhr.sd_acc}\,\mathrm{m/s^2}$ and $\kappa_\omega=\pv[2]{stats.mhr.sd_gyro}\,\mathrm{rad/s}$ are the root-mean-square values per axis of canonicalized joint-center signals on \pv[0]{stats.mhr.n_clips} training clips.
The reference channels thus carry the segment tilt, but nothing about segments without a device. Their horizontal part depends on the forward axis of $W_d$, which gait fixes only up to its sign, so this part is randomized in training (\cref{sec:stage1}).

\paragraph{Walking frame.}
As a training-free alternative to the calibrator (\cref{sec:context}), the Walking frame~\cite{mizell2003gravity,kunze2009facing} estimates the up axis and a sign-ambiguous forward axis in device axes, as for $W_d$, from the context windows that pass a gait test on the periodicity of their acceleration magnitude and leaves the reference channels at zero if none does. %
Training simulates it from $W_d$ with the measured spread of its up axis. %

\subsection{Encoder and motion pretraining}
\label{sec:stage1}

\paragraph{Encoder.} The graph encoder $E_\theta$ (\cref{fig:method}, top right) is a spatio-temporal graph network over the \pvc{22} sites with the bones as edges, as in ST-GCN~\cite{yan2018stgcn} and AnyMo~\cite{chen2026anymo}, but over tokens of \pvc{0.2}-s patches rather than samples. %
A worn site holds one token per patch, a linear projection of its patch of $x_d$. An unworn site holds a learned mask token, which relays information between worn sites. %
Every token also receives learned embeddings of its site, time and gyroscope flag and a linear map of the \pvc{44} worn-site and gyroscope flags. %
Eight residual graph blocks each update every site from itself, its parent and its children with learned edge weights and then convolve along time over nine patches with stride one. %
The tokens of the worn sites, averaged over these sites and over all time steps that hold signal, projected and normalized, give the window embedding $v$. Tokens have width \pvc{256} and $v$ has \pvc{512} dimensions. %

\paragraph{Stage~1: Motion pretraining.} The encoder learns to predict how a movement appears under a given setup, with the reference as the only cue to the mounting of the target device (\cref{fig:method}, bottom left).
Each batch simulates $N=\pvc{64}$ movements under four views, namely a random set of \pvc{1} to \pvc{5} sites, the same set re-worn (\cref{sec:virtual}), a second random set and all \pvc{22} sites. %
We adopt the predictor and contrastive loss of AnyMo~\cite{chen2026anymo}, which predicts the pooled features of a second view, but let the predictor target one device and condition it on that device's setup reference. %
Each step also simulates \pvc{32} further movements on the same devices as calibrator context. The calibrator estimates one $\hat Q_d$ per device from a random subset of \pvc{6} to \pvc{32} of them. %
The reference of device $d$ in view $\beta$ is this estimate rotated about the up axis of $W_d$ by an angle $\eta$, uniform except at the pelvis, legs and lower spine, where gait fixes the forward axis up to its sign and $\eta$ is drawn from $\{0,\pi\}$. This keeps the encoder from relying on a forward direction the calibrator cannot estimate. The reference replaces $\hat Q_d$ in \cref{eq:input} and is zeroed for \pvc{20}\,\% of the devices so that the encoder also works without one. %
For a source view $\alpha$, a target view $\beta\neq\alpha$ and a target device $d$ of $\beta$, the predictor $P_\xi$, a six-block Transformer over the \pvc{50} patches, takes the source features of movement $i$ averaged over the worn sites $\mathcal S_\alpha$, with a query added at every patch that sums an embedding $e_d$ of the site and an MLP of the first two columns of the rotated estimate, %
\begin{equation}
\hat z_i=P_\xi\Big(\frac{1}{|\mathcal S_\alpha|}\sum_{s\in\mathcal S_\alpha}E_\theta(X_i^\alpha)_s,\ e_d+\mathrm{MLP}\big(R_y(\eta)\,\hat Q_d\big)\Big),
\label{eq:pred}
\end{equation}
where $E_\theta(X)_s$ is the token sequence of site $s$ for the input $X$.
The predicted sequence $\hat z_i$ must find its movement among the $N$ candidates $z_j=\mathrm{sg}\big[E_\theta(X_j^\beta)_d\big]$, the target features under a stop-gradient $\mathrm{sg}$, by InfoNCE~\cite{oord2018cpc},
\begin{equation}
\mathcal L_1=-\frac{1}{|\mathcal G|\,N}\sum_{(\alpha,\beta,d)\in\mathcal G}\,\sum_{i=1}^{N}\log\frac{\exp(\rho_{ii}/\tau)}{\sum_{j=1}^{N}\exp(\rho_{ij}/\tau)},
\label{eq:stage1}
\end{equation}
where $\rho_{ij}$ is the cosine between $\hat z_i$ and $z_j$ averaged over the patches valid in both, $\mathcal G$ holds the triples $(\alpha,\beta,d)$ with up to \pvc{3} target sites per view pair and $\tau=\pvc{0.1}$. %
Stage~1 minimizes $\mathcal L_1+\mathcal L_{\mathrm{cal}}$ over encoder, predictor and calibrator. %

\subsection{Captions and caption alignment}
\label{sec:annotation}

Stage~2 aligns the encoder with text, using Nymeria's narrations (\pv[0]{data.d3.train.human} lines on the training windows) and captions generated from the measured motion of each of these windows. %
To generate captions, a rule-based program measures the MHR skeleton in a pelvis-centered frame and produces a time-stamped digest of states and changes, from which an LLM (Qwen3.6-27B~\cite{qwen2026qwen36}) writes four captions per window, instructed not to invent objects, scenery or movements.
Because the digest carries the motion but not its purpose, the generated captions describe kinematics rather than activities. %
For one window the generated caption reads ``a person bends forward deeply, shifts weight backward and forward, and rotates the torso'', the narration ``picking up another decorative item on the table''.
In every step, a window is paired with one of its narrations, if it has any, with probability \pvc{0.5} and with one of its generated captions otherwise. %
The narrations-only and generated-only models of \cref{tab:context} set this probability to one and zero. %

\label{sec:stage2}%
\paragraph{Stage~2: Caption alignment.} The encoder then learns to match captions (\cref{fig:method}, bottom middle).
It embeds window $i$ as $v_i$, while the CLIP ViT-B/32 text encoder~\cite{radford2021clip}, as in UniMTS but frozen, embeds its caption as $u_i$.
With a learned logit scale $\gamma$ and a batch of $B$ windows, the symmetric contrastive loss of CLIP reads
\begin{equation}
\begin{aligned}
\mathcal L_2={}&-\frac{1}{2B}\sum_{i=1}^{B}\Big[\log\frac{e^{\gamma\,v_i^{\top}u_i}}{\sum_{j}e^{\gamma\,v_i^{\top}u_j}}\\
&+\log\frac{e^{\gamma\,u_i^{\top}v_i}}{\sum_{j}e^{\gamma\,u_i^{\top}v_j}}\Big],
\end{aligned}
\label{eq:stage2}
\end{equation}
and stage~2 minimizes $\mathcal L_2+\lambda_A\mathcal L_A$, where the auxiliary loss $\mathcal L_A$, weighted by $\lambda_A=\pvc{0.1}$, keeps the motion of every worn joint recoverable from the encoder's tokens as the mean of $\big(h_A(z_{s,p})-y_{s,p}\big)^2$ over the entries of worn sites $s$ and valid patches $p$, with gyroscope entries only where a gyroscope is worn. %
Here $h_A$ is a linear head on the token $z_{s,p}$. The target $y_{s,p}$ holds the noise-free specific force and angular velocity of the patch at the joint center, aligned with the true gravity direction and divided by $\kappa_f$ and $\kappa_\omega$.
Because the calibrator's estimates are imperfect, stage~2 trains under its errors. After stage~1, an error bank stores them once as the rotations $\Delta=M_b^{\top}\hat Q_b^{\top}W_b$ from the true mounting $M_b$ of a simulated device $b$ at the same site to the mounting $\hat Q_b^{\top}W_b$ that the frozen calibrator implies for it, for \pv[0]{bank.tw.devices_per_site} devices per site and gyroscope condition, each from six held-out Nymeria windows. %
The median error with a gyroscope is \pv[1]{bank.tw.gyro.up.median}$^\circ$ for the estimated up axis and \pv[0]{bank.tw.gyro.b.median}$^\circ$ for the reference channels, whose horizontal part also depends on the forward sign that gait leaves open. %
Stage~2 draws $M_d$ uniformly from $SO(\pvc{3})$, covering every mounting. For every window and site, one error is drawn uniformly from the bank entries of that site and gyroscope condition and applied to its mounting, so that the calibrator's estimate in \cref{eq:input} is replaced by %
\begin{equation}
\hat Q_d=W_d\,(M_d\Delta)^{\top},\qquad b_d(t)=\hat Q_d\,\hat g_d(t),
\label{eq:bank}
\end{equation}
which equals the estimate $\hat Q_b$ the calibrator made for device $b$ whenever $M_d=M_b$.
Half of the stage-2 windows are accelerometer-only at all sites.

\subsection{Inference with class names and anchors}
\label{sec:test}

At test, all windows of one participant at one site are treated as one device, assumed to share one mounting (\cref{fig:method}, bottom right). %
From twelve unlabeled windows $\mathcal K$ of the device, the frozen calibrator repeatedly drops the window whose removal leaves the smallest tilt uncertainty, $\mathcal K\leftarrow\mathcal K\setminus\{\arg\min_{n\in\mathcal K}\sigma_\uparrow(\mathcal K\setminus\{n\})\}$, until six remain and then estimates $\hat Q_d$ from them (\cref{tab:main}, last row). %
Every scored window of the device is encoded with $b_d(t)=\hat Q_d\,\hat g_d(t)$ (\cref{eq:input}).
Zero-shot classification assigns each window to the class whose text embedding has the highest cosine similarity to $v$, where the class text, embedded by each model's own frozen text encoder, is either the class name as given in the dataset or an anchor.

An anchor replaces the class-name embedding with an average of text embeddings, as in CLIP's prompt ensembling~\cite{radford2021clip} with descriptions~\cite{pratt2023cupl,menon2023description}. It uses the class name of the target dataset but no labeled target window. %
Let $n_c$ be the embedding of the class name, $r_c$ the mean embedding of the ten training narrations nearest to the name in all-mpnet space~\cite{song2020mpnet,sbert2021allmpnet} and $g_{cj}$ the mean embedding of ten descriptions of a person performing the activity, written by an LLM in sampling round $j\in\{\pvc{1},\pvc{2},\pvc{3}\}$.
With $U$ the unit normalization, the anchor of class $c$ weighs the name and its nearest narrations equally against the LLM-written descriptions, %
\begin{equation}
a_c = U\Big(U\big(n_c+U(r_c)\big) + U\Big(\tfrac{\pvc{1}}{\pvc{3}}\sum_{j} U(g_{cj})\Big)\Big),
\label{eq:anchor}
\end{equation}
and each window is assigned the class $\hat c=\arg\max_c v^{\top}a_c$. %

\subsection{Language model on the encoder}
\label{sec:lmbridge}

For questions a class name cannot answer (\cref{sec:intro}), we also train a language model (LM) on top of the encoder.
An MLP projector maps the features of the frozen SetCon encoder, computed with the calibrator's estimate of the setup reference, to one token for the worn-site flags and one token per worn site and time patch, which precede a Qwen3.5-0.8B~\cite{qwen2026qwen35} prompt that lists the worn sites. %
The projector is first trained to caption virtual windows with the LM frozen, then jointly with the LM on an equal mix of captions and question-answer pairs that an LLM writes from the narrations and digest of each window.
Both stages minimize the next-token loss of the answer, with contrastive and margin terms that tie the answer to the IMU tokens. %

\subsection{Implementation details}
\label{sec:impl}

The encoder has \pv{impl.enc.params.m}\,M parameters and the calibrator \pv[2]{impl.cal.params.m}\,M. Every stage uses AdamW.
Stage~1 trains for \pv[0]{impl.s1.epochs} epochs at constant learning rates of $\pv[0]{impl.s1.lr.e4}\cdot10^{-4}$ for encoder and predictor and $\pv[0]{impl.s1.callr.e5}\cdot10^{-5}$ for the calibrator. %
Stage~2 trains the encoder with its new projection and heads for \pv[0]{data.s2.epoch.read} epochs on batches of $B=\pv[0]{impl.s2.batch}$ windows at a learning rate of $\pv[0]{impl.s2.lr.e4}\cdot10^{-4}$ with weight decay \pv[2]{impl.s2.wd} and a linear warm-up over the first \pv[0]{impl.s2.warm} epochs. We evaluate the exponential moving average of its weights (decay \pv[3]{impl.s2.ema}). This schedule is the same for every model, variant and seed. %
Training and evaluating one model takes at most \pv{compute.one.h}~GPU-hours on one NVIDIA H100. %

%% file: sec/4_exp_a.tex
\section{Experiments}\label{sec:exp}

\subsection{Experimental protocol}\label{sec:setup}
\input{floats/tab1}

SetCon is trained only on virtual IMU from Nymeria~\cite{ma2024nymeria} (\cref{sec:virtual}), with every recording of twelve participants held out for validation. It is tested on the \pv[0]{set.m13.n} real datasets of the UniMTS suite~\cite{zhang2024unimts} that record a gyroscope on every device. %
Mean-13 (\cref{sec:intro}) averages the accuracy on the scored windows of each dataset, with macro-F1 averaged in the same way. %
The baselines are UniMTS~\cite{zhang2024unimts} and AnyMo~\cite{chen2026anymo} with their released weights. UniMTS uses the accelerometer of one window, while AnyMo uses accelerometer and gyroscope with the body segment in its prompt. %
For its reference, SetCon also uses unlabeled context windows of the same device, never scored.
In the last row of \cref{tab:main} the calibrator chooses six of twelve such windows (\cref{sec:test}) and every other result uses six random ones unless stated otherwise, averaged over five draws.

\subsection{Main results}\label{sec:main}

With six of twelve unlabeled windows per device as context, SetCon reaches \pv[1]{mix.psel6.w12.nm.acc.m13}\,\% Mean-13 accuracy with class names and \pv[1]{mix.psel6.w12.ac.acc.m13}\,\% with anchors (\cref{tab:main}), whereas AnyMo, the stronger baseline, reaches \pv[1]{anymo.window.w12.nm.acc.m13}\,\% and \pv[1]{anymo.window.w12.ac.acc.m13}\,\% on the same windows, a relative gain of \pv[0]{r.mix-anymo.psel6.w12.nm.acc.m13}\,\% and \pv[0]{r.mix-anymo.psel6.w12.ac.acc.m13}\,\%.
The lead holds in macro-F1 and with random context windows. With six random windows, SetCon reaches \pv[1]{mix.own.p1.nm.acc.m13}\,\% accuracy and \pv[1]{mix.own.p1.nm.f1.m13}\,\% macro-F1 with class names, compared with \pv[1]{anymo.window.p1.nm.acc.m13}\,\% and \pv[1]{anymo.window.p1.nm.f1.m13}\,\% for AnyMo. %

SetCon's embeddings also separate the activities best. A class-mean readout tests this without class names, which takes the wording of the classes out of the comparison. It averages the embeddings of the labeled windows of each class and assigns every window to the class with the nearest mean. With class means computed from all scored test windows, it reaches \pv[1]{mix.own.p1.nm.sep.m13}\,\% Mean-13 accuracy for SetCon, compared with \pv[1]{anymo.window.p1.nm.sep.m13}\,\% for AnyMo and \pv[1]{unimts.window.p1.nm.sep.m13}\,\% for UniMTS.
SetCon is therefore ahead of both baselines even when no class name is used. %

\subsection{Analysis of the setup reference}\label{sec:context}

\input{floats/tab2}
SetCon relies on its setup reference (\cref{tab:context}).
Zeroing it at test lowers Mean-13 accuracy from \pv[1]{mix.own.p1.nm.acc.m13}\,\% to \pv[1]{mix.zero.p1.nm.acc.m13}\,\% with class names and from \pv[1]{mix.own.p1.ac.acc.m13}\,\% to \pv[1]{mix.zero.p1.ac.acc.m13}\,\% with anchors. A model trained without a reference lies \pv[1]{d.mix-wo.own.p1.nm.acc.m13@abs} and \pv[1]{d.mix-wo.own.p1.ac.acc.m13@abs} points below SetCon. %
One context window already restores most of the drop (\pv[1]{mix.ks1.p1.nm.acc.m13}\,\% with names). The training-free Walking frame of \cref{sec:reference} comes close to the calibrator, with \pv[1]{mix.walk.p1.nm.acc.m13}\,\% using the same weights and \pv[1]{walk.walk.p1.nm.acc.m13}\,\% when the encoder is also trained with it, so the encoder relies on the setup reference itself rather than on one estimator of it. %
The learned calibrator matters where the context windows contain no gait and the Walking frame leaves the reference at zero. On these devices (\pv[0]{main13.nogait.share}\,\% of the scored windows), it outperforms the zeroed reference by \pv[1]{main13.nogait.gain.nm} points with names. %

\input{floats/tab_groups}
Built for selective invariance, SetCon leads where full invariance fails (\cref{tab:groups}). %
With names it reaches a mean class accuracy of \pv[1]{err.grp.posture.mix.nm}\,\% on postures against \pv[1]{err.grp.posture.anymo.nm}\,\% for AnyMo and \pv[1]{err.grp.dirvert.mix.nm}\,\% on stairs and elevators against \pv[1]{err.grp.dirvert.anymo.nm}\,\%. Both groups depend on the direction of gravity, which full invariance removes together with the mounting. %
Trained on the same data without a reference, the encoder reaches only \pv[1]{err.grp.posture.wo.nm}\,\% on postures, close to AnyMo and below SetCon, while no other group of \cref{tab:groups} changes by more than \pv[1]{err.grp.nonposture.mix-wo.nm.maxabs} points. On UCI-HAR~\cite{anguita2013ucihar}, its embeddings mix lying with sitting as AnyMo's do, while SetCon's keep them apart (\cref{fig:pca}). %
With the reference zeroed, SetCon fails like the baselines in \cref{sec:intro} and classifies, with anchors, on average \pv[0]{post.mixzero.ac.lie}\,\% of the sitting, standing and lying windows of the same datasets as lying. %

%% file: floats/tab1.tex
\input{tables/tab1_main_float}

%% file: tables/tab1_main_float.tex
\begin{table*}[t]
\centering
\caption{\textbf{Zero-shot classification.} Mean-13 accuracy and macro-F1 (\%) with class names and anchors, plus the Mean-13 accuracy of the class-mean readout (\cref{sec:main}). The calibrator estimates the reference from unlabeled context windows of each device, never scored. In the last row it chooses six of twelve context windows. The first three rows share their scored windows. On the last row's scored windows, UniMTS reaches \pv[1]{unimts.window.w12.nm.acc.m13} and \pv[1]{unimts.window.w12.ac.acc.m13} and AnyMo \pv[1]{anymo.window.w12.nm.acc.m13} and \pv[1]{anymo.window.w12.ac.acc.m13} with names and anchors. SetCon rows give mean $\pm$ sd over three seeds.}%
\label{tab:main}
\input{tables/tab1_main}
\end{table*}

%% file: tables/tab1_main.tex
\ifdefined\tbpm\else\gdef\tbpm#1#2{#1\,$\pm$\,#2}\fi
\ifdefined\tbna\else\gdef\tbna{--}\fi
\ifdefined\tbbox\else\newsavebox{\tbbox}\fi
\begingroup
\small\setlength{\tabcolsep}{3.2pt}
\newif\iftbinpm
\def\tbpm#1#2{{\tbinpmtrue#1\,$\pm$\,#2}}%
\DeclareRobustCommand\pv[2][1]{\csname pv@show\endcsname{#2}{#1}\iftbinpm\else\phantom{\,$\pm$\,0.0}\fi}%
\sbox{\tbbox}{%
\begin{tabular}{@{}lllccccc@{}} %
\toprule
& & & Class-mean & \multicolumn{2}{c}{Mean-13 acc.} & \multicolumn{2}{c}{Mean-13 macro-F1} \\
\cmidrule(lr){4-4}\cmidrule(lr){5-6}\cmidrule(lr){7-8}
Method & Gyro. & Context & readout & names & anchors & names & anchors \\
\midrule
\csname @@input\endcsname tables/tab1_main_rows\pvsuffix.tex 
\bottomrule
\end{tabular}}%
\ifdim\wd\tbbox>\linewidth\resizebox{\linewidth}{!}{\usebox{\tbbox}}\else\usebox{\tbbox}\fi
\endgroup

%% file: floats/tab2.tex
\input{tables/tab2_context_float}

%% file: tables/tab2_context_float.tex
\begin{table}[t]
\centering
\caption{\textbf{Ablations.} Mean-13 accuracy (\%) with six random context windows (\cref{tab:main}, third row). Mean $\pm$ sd over three seeds. Free answers are judged against the class names (\cref{sec:lm}).}%
\label{tab:context}%
\input{tables/tab2_context}
\end{table}

%% file: tables/tab2_context.tex
\ifdefined\tbpm\else\gdef\tbpm#1#2{#1\,$\pm$\,#2}\fi
\ifdefined\tbna\else\gdef\tbna{--}\fi
\ifdefined\tbbox\else\newsavebox{\tbbox}\fi
\begingroup
\small\setlength{\tabcolsep}{6pt}
\newif\iftbinpm
\def\tbpm#1#2{{\tbinpmtrue#1\,$\pm$\,#2}}%
\DeclareRobustCommand\pv[2][1]{\csname pv@show\endcsname{#2}{#1}\iftbinpm\else\phantom{\,$\pm$\,0.0}\fi}%
\sbox{\tbbox}{%
\begin{tabular}{@{}lcc@{}}
\toprule
& \multicolumn{2}{c}{Mean-13} \\
\cmidrule(lr){2-3}
Variant & names & anchors \\ %
\midrule
\multicolumn{3}{@{}l}{\emph{Reference at test}}\\ %
\csname @@input\endcsname tables/tab2_context_rows\pvsuffix.tex 
\bottomrule
\end{tabular}}%
\ifdim\wd\tbbox>\linewidth\resizebox{\linewidth}{!}{\usebox{\tbbox}}\else\usebox{\tbbox}\fi
\endgroup

%% file: floats/tab_groups.tex
\input{tables/tab_groups_float}

%% file: tables/tab_groups_float.tex
\begin{table}[t]
\centering
\caption{\textbf{Accuracy per motion group.} Mean class accuracy (\%) with class names over Mean-13 classes, grouped by name. SetCon uses six random context windows (three seeds, mean $\pm$ sd).}%
\label{tab:groups}
\input{tables/tab_groups}
\end{table}

%% file: tables/tab_groups.tex
\providecommand{\tbq}[2]{#1$\pm$#2}%
\begingroup
\footnotesize\setlength{\tabcolsep}{7pt}
\begin{tabular}{@{}lccc@{}}
\toprule
Group (no.\ of classes) & UniMTS & AnyMo & SetCon (ours) \\ %
\midrule
\csname @@input\endcsname tables/tab_groups_names_rows.tex 
\bottomrule
\end{tabular}
\endgroup

%% file: sec/4_lang.tex
\subsection{Language access}
\label{sec:language}
\input{floats/fig_pca}
Training captions must describe activities. Aligned with generated captions alone, which describe kinematics rather than activities, the encoder reaches \pv[1]{gen.own.p1.nm.acc.m13}\,\% Mean-13 accuracy with class names, \pv[1]{d.mix-gen.own.p1.nm.acc.m13@abs} points below SetCon. %
Trained on narrations alone, it reaches \pv[1]{nar.own.p1.nm.acc.m13}\,\% with names, above SetCon's \pv[1]{mix.own.p1.nm.acc.m13}\,\%, but \pv[1]{nar.own.p1.ac.acc.m13}\,\% with anchors, below its \pv[1]{mix.own.p1.ac.acc.m13}\,\%. Generated captions trade a little class-name accuracy for a larger gain from anchors, which grows with their share from \pv[1]{d.nar:own:ac-nar:own:nm.p1.acc.m13} points without them to \pv[1]{d.gen:own:ac-gen:own:nm.p1.acc.m13} with them alone. %
The growth comes mainly from the descriptions half of the anchor in \cref{eq:anchor}. The LLM-written descriptions, like the generated captions, say how the body moves, whereas the narrations in the other half describe activities, as class names do. Added to the class name, the descriptions bring \pv[1]{anc.split.nar.name_desc-nm.m13}, \pv[1]{anc.split.mix.name_desc-nm.m13} and \pv[1]{anc.split.gen.name_desc-nm.m13} points for the narrations-only model, SetCon and the generated-only model. Zero-shot classification thus works best with prompts that resemble the training text. %

Class-name prompts recover only part of the separation in the embeddings. SetCon stays \pv[0]{diag.rdnm.mix.m13@abs} points below the class-mean readout. %
Anchors, which use no labeled window, close part of the gap, adding \pv[1]{d.mix:own:ac-mix:own:nm.p1.acc.m13@abs} points for SetCon, compared with \pv[1]{d.anymo:window:ac-anymo:window:nm.p1.acc.m13@abs} for AnyMo and \pv[1]{d.unimts:window:ac-unimts:window:nm.p1.acc.m13@abs} for UniMTS (\cref{tab:main}). %

%% file: floats/fig_pca.tex
\begin{figure}[t]
\centering
\includegraphics[width=\linewidth]{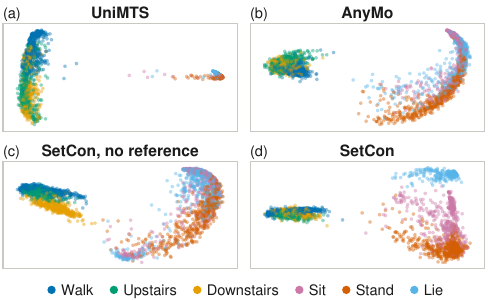}
\caption{\textbf{Embeddings of UCI-HAR.} Each panel shows the first two principal components of one model's embeddings of the scored windows, with the first seed in (c) and (d). Lying mixes with sitting in (b) and (c) but forms its own cluster in (d).}%
\label{fig:pca}
\end{figure}

%% file: sec/4_exp_b.tex
\label{sec:lm}%
The language model of \cref{sec:lmbridge}, trained on each of the three SetCon seeds, answers ``What is the person doing?'' in free text. An LLM judge~\cite{qwen2026qwen36} maps each answer to a class name of the dataset or to none. %
The language model reaches \pv[1]{lm.full.free.m13}\,\% Mean-13 accuracy (\cref{tab:context}), in every seed above the \pv[1]{mix.own.p1.nm.acc.m13}\,\% of the encoder itself with class-name prompts. Vision-language models, in contrast, classify images worse than their image encoder~\cite{zhang2024vlmclassify}. %
With anchors, however, the same encoder reaches \pv[1]{mix.own.p1.ac.acc.m13}\,\% and is more accurate in every seed. All three stay below the \pv[1]{mix.own.p1.nm.sep.m13}\,\% of the class-mean readout (\cref{tab:main}). %
For a closed set of classes, cosine similarity to the anchors thus remains the most accurate and efficient zero-shot classifier, while the language model handles questions a class prompt cannot express. On held-out virtual Nymeria windows with candidate answers, it answers five-way questions on execution, order and body region correctly in \pv[0]{lm.qa.execution.own}, \pv[0]{lm.qa.order.own} and \pv[0]{lm.qa.body.own}\,\% of cases (\pv[0]{lm.qa.body.chance}\,\% chance) and selects the correct activity of four in \pv[0]{lm.qa.mc.own}\,\% (\pv[0]{lm.qa.mc.chance}\,\% chance).
It also answers the five-way activity question written by the LLM correctly in \pv[0]{lm.qa.activity.own}\,\% of cases (\pv[0]{lm.qa.activity.chance}\,\% chance) and yes/no questions in \pv[0]{lm.qa.yesno.own}\,\% (\pv[0]{lm.qa.yesno.chance}\,\% chance).
Its answers depend on the signal. On a subset of the windows of \cref{tab:context}, they reach \pv[1]{lm.ev.free.noimu.m13}\,\% Mean-13 accuracy without the IMU tokens. Given the IMU tokens of another window, they give that window's class in \pv[0]{lm.ev.free.donor.follow.m13}\,\% of cases. %

%% file: sec/5_conclusion.tex
\section{Conclusion}\label{sec:conclusion}

For IMU-language learning, the sensor setup is a source of information.
An encoder invariant to how a device is worn loses the direction of gravity relative to the body, which separates postures. With selective invariance, SetCon retains it through the setup reference, which a calibrator estimates from unlabeled windows.
Trained only on Nymeria, it outperforms UniMTS and AnyMo on average over \pv[0]{set.m13.n} real datasets on which all three models receive their full input.
Zero-shot accuracy also depends on the class prompt. Anchors bring \pv[1]{d.mix:own:ac-mix:own:nm.p1.acc.m13} points for SetCon, compared with \pv[1]{d.nar:own:ac-nar:own:nm.p1.acc.m13} for an encoder trained without generated captions. With anchors, the encoder is more accurate than a language model built on it (\pv[1]{lm.full.enc.ac.m13} against \pv[1]{lm.full.free.m13}\,\%). %
3D human motion does more than provide training data. Real IMU datasets rarely record the mounting, so the calibrator's target comes from simulation. To learn the setup at scale, general-purpose IMU-language models therefore require 3D human motion. %

\paragraph{Limitations.}
SetCon assumes that the body site of a device is known and that its mounting stays fixed while the device is worn.
The \pv[0]{set.m13.n} test sets are public benchmarks with known body sites. Unscripted daily life, in which devices slip and sites are unknown, remains to be tested. %

\paragraph{Impact.}
With the best-separated embeddings in \cref{tab:main}, SetCon can serve as a general-purpose IMU encoder for the phones, watches and glasses people already wear.
Recognizing activities on any wearable without labels could lower the cost of health and rehabilitation monitoring.
Beyond classification, its embeddings can support retrieval, captioning and question answering over long recordings.
In egocentric multimodal models, it adds to vision and audio a signal that follows the body out of the camera's view.
AI agents and robots could use SetCon to perceive what the people they cooperate with are doing.
Estimating the setup may also help other body-worn sensors, such as cameras.
SetCon learns from IMU simulated on 3D body motion, which motion capture and human mesh recovery from video both produce.
Progress in 3D human motion therefore becomes training data for wearable sensing.
Because SetCon reveals what wearers do, deployment requires their informed consent.